\documentclass[final,5p,times,twocolumn,article]{elsarticle}
\usepackage{lineno}
\usepackage[hidelinks]{hyperref}
\usepackage{algorithm}
\usepackage{algpseudocode}
\usepackage{graphicx}
\usepackage{amsmath}   
\usepackage{booktabs}
\usepackage{multirow}
\usepackage{amssymb}
\usepackage{mathptmx}
\usepackage{textcomp}
\usepackage{hyperref}

\hypersetup{
    pdftitle={EP-SAM: Weakly Supervised Histopathology Segmentation via Enhanced Prompt with Segment Anything}
}

\def\pmbanner{{\hrule height 1 pt}\vskip35pt{}\vskip35pt{\hrule height 4pt}\vskip20pt}

\begin{document}

\begin{frontmatter}

\title{\pmbanner EP-SAM: Weakly Supervised Histopathology Segmentation via Enhanced Prompt with Segment Anything}

\author[1,2]{Joonhyeon Song\fnref{fn1}}
\ead{thdwnsgus0706@gmail.com}

\author[1]{Seohwan Yun\fnref{fn1}}
\ead{dbstjghks@naver.com}

\author[1]{Seongho Yoon}

\author[1]{Joohyeok Kim}

\author[1]{Sangmin Lee\corref{cor1}}
\ead{smlee5679@kw.ac.kr}

\cortext[cor1]{Corresponding author}
\fntext[fn1]{These authors contributed equally to this work.}

\affiliation[1]{organization={Information Convergence, Kwangwoon University}, 
                city={Seoul}, 
                country={South Korea}}
\affiliation[2]{organization={QI LAB},
                city={Seoul},
                country={South Korea}}

\begin{abstract}
 This work proposes a novel approach beyond supervised learning for effective pathological image analysis, addressing the challenge of limited robust labeled data. Pathological diagnosis of diseases like cancer has conventionally relied on the evaluation of morphological features by physicians and pathologists. However, recent advancements in compute-aided diagnosis (CAD) systems are gaining significant attention as diagnostic support tools. Although the advancement of deep learning has improved CAD significantly, segmentation models typically require large pixel-level annotated dataset, and such labeling is expensive. Existing studies not based on supervised approaches still struggle with limited generalization, and no practical approach has emerged yet. To address this issue, we present a weakly supervised semantic segmentation (WSSS) model by combining class activation map and Segment Anything Model (SAM)-based pseudo-labeling. For effective pretraining, we adopt the SAM—a foundation model that is pretrained on large datasets and operates in zero-shot configurations using only coarse prompts. The proposed approach transfer enhanced Attention Dropout Layer's knowledge to SAM, thereby generating pseudo-labels. To demonstrate the superiority of the proposed method, experimental studies are conducted on histopathological breast cancer datasets. The proposed method outperformed other State-of-the-Art (SOTA) WSSS methods across three datasets, demonstrating its efficiency by achieving this with only 12GB of GPU memory during training. Our code is available at : \href{https://github.com/QI-NemoSong/EP-SAM}{https://github.com/QI-NemoSong/EP-SAM}.
\end{abstract}


\end{frontmatter}


\section{Introduction}
\label{sec1}

Cancer is one of the most critical diseases in the world, posing a significant risk to people’s health due to its high mortality rate \cite{siegel2024cancer}. Accurate diagnosis is crucial for effective treatment and management. Currently, the `gold standard' for identifying and quantifying cancer involves histopathological analysis of tissue biopsies. This method relies on the visual assessment by pathologists and clinicians. Several computer-aided diagnosis (CAD) systems based on machine learning methods have been developed to alleviate the burden on experts. With recent advances in deep learning, CAD has been applied with remarkable performance in several tasks in this field, including image classification, object localization, and semantic segmentation. Semantic segmentation, which aims to extract a region of interest from each patch, plays an important role in distilling informative morphological attributes for professionals. Segmentation performance has been enhanced with the inception of state-of-the-art (SOTA) methods utilizing convolutional neural networks and vision transformer (ViT) backbones \cite{ronneberger2015u, xie2021segformer, chen2017deeplab, chen2017rethinking, zhang2022segvit, strudel2021segmenter}. However, these methods require large amounts of pixel-level annotated data for training. Obtaining such datasets is often time-consuming and expensive, especially in the histopathology domain, due to the need for skilled domain expertise in labeling.

 Weakly supervised semantic segmentation (WSSS), which uses coarse-grained annotated data such as points and bounding boxes for supervision has emerged as an alternative approach. Numerous WSSS methods have been proposed in the medical field \cite{lin2023nuclei, srinidhi2022self, pati2023weakly}. Recently, WSSS methods that use less costly image-level labels have gained significant attention and achieved remarkable results. Conventional algorithms are predominantly based on class activation map (CAM) and are typically divided into two phases, with the first being the generation of pseudo-labels using a classifier and CAM, followed by the optimization of refined masks through post-processing methods such as dense conditional random field (DenseCRF \cite{liang2015semantic}), and the second being the training of these refined pseudo-labels using an off-the-shelf segmenter.

However, CAM suffers from some well-known problems, such as false activation and partial activation \cite{gao2021ts}, which limits it from detecting the boundaries of objects accurately. These challenges are particularly amplified in histopathological images that feature more blurred \cite{gu2024lesam} and homogeneous boundaries \cite{fang2023weakly} than natural images. Moreover, the performance of WSSS is bounded above by the performance of a model with fully supervised training on pixel-level annotation data, depending on the capability of the off-the-shelf segmenter.

Although, SAM, a foundation model pre-trained on large-scale data, exhibits remarkable performance. It far surpasses conventional segmentation models, even in zero-shot learning scenarios, by utilizing prompts during inference. Numerous recent studies have attempted to utilize SAM in the medical field, demonstrating its potential \cite{mazurowski2023segment, huang2024segment, zhang2024segment, zhang2023customized, zhu2024medical}, yet certain drawbacks remain unresolved.
First, significant performance variance is observed in segmentation masks depending on the prompts \cite{kirillov2023segment}. Second, owing to the domain gap between natural and medical images \cite{gu2024lesam}, the zero-shot performance is notably inferior in the latter case. 

 In this context, the essential research problem is, 
\\\textit{How can we leverage SAM's performance efficiently in weakly supervised histopathology segmentation scenarios without having to input additional prompts from the ground truth?}\\
We propose the weakly supervised pseudo-labeling method to address this problem. 

The main contributions of our paper are as follows:

1. We have enhanced the attention dropout layer (ADL) by incorporating explicit visual prompting, which mitigates incompleteness issues such as partial and false activations inherent in CAM-based approaches. Our experiments on various breast cancer datasets demonstrate that the enhanced module outperforms existing CAM-based alternatives in terms of generating the initial pseudo-labels. To the best of our knowledge, this study represents the first application of explicit visual prompting in CAM-based methods.

2. We have devised a framework that optimizes SAM performance in weakly supervised breast cancer segmentation without relying on ground-truth based prompts. Our approach outperforms current WSSS SOTA methods and several fully supervised methods.

3. Our approach includes a SAM fine-tuning stage; but, it has been designed in a memory-efficient manner by fine-tuning only the lightweight decoder. This design choice reduces the computational requirements significantly while maintaining high performance and allows our framework to operate with only 12 GB of GPU memory.

\section{Related Work}
\label{sec2}
\subsection{Explicit Visual Prompting in Computer Vision}
Explicit visual prompts extracted from input images have been used to guide models to focus on specific content during training \cite{liu2023explicit, chen2023sam}. In particular, by leveraging high-frequency components, these approaches exhibit remarkable performance in tasks where distinguishing between the foreground and background is challenging, e.g., camouflaged object detection and shadow detection. However, these studies primarily focused on parameter-efficient fine-tuning \cite{liu2022few} with the aim of enabling efficient learning with fewer parameters. Other than segmentation, research using explicit content extracted from input data to guide intended learning outcomes in other fields remains limited. Inspired by these, we focused on resolving the partial activation problem in CAM-based methods, especially in medical datasets where distinguishing foreground from background is still challenging.

\subsection{WSSS in Histopathology}

Obtaining detailed annotations for medical images is challenging and requires specialized expertise. To address this issue, multiple-instance learning (MIL) has been adapted for WSSS in medical imaging. For instance, Xu et al.  \cite{xu2012multiple} introduced a multiple clustered instance learning framework called CAMEL to differentiate between cancerous and non-cancerous areas. It treats histopathological images as bags and subdivided patches as instances. Jia et al. \cite{jia2017constrained} developed DWSMIL to identify cancerous regions in histopathological images. Some alternatives to MIL have also been proposed. Han et al. \cite{han2022multi} devised progressive drop out attention and classification gate mechanism for WSSS with H\&E stained images. The aforementioned approaches yielded significant results; however, they remain suboptimal owing to their poor generalizability across various datasets. Further research is required to yield a dominant method for this purpose.

\subsection{Effective Prompts for SAM}

SAM utilizes various prompt types, such as masks, bounding boxes, and points, with performance varying significantly in medical images where  foreground and background distinction is often unclear. Among the various aforementioned types of prompts, using masks directly has been demonstrated to yield poor performance \cite{kweon2024sam}, whereas the universal utilization of bounding boxes is challenging, especially in sparsely annotated data, where using entire boxes is not ideal. Consequently, we choose to use point-type prompts for the seeds. In this work, to generate better seeds, we propose a seed-prompting module based on pixel-level entropy. 

\subsection{Transferring Knowledge to SAM for WSSS}

Numerous studies have focused on developing  effective WSSS methods by incorporating greedy algorithms with  SAM. For instance, \cite{ren2024grounded} utilized the Ground DINO Object Detection method \cite{liu2023grounding}, to generate bounding boxes, which were then used as prompts for SAM, whereas Yang et al. \cite{yang2024foundation} generated seeds using CLIP. These studies reported methods to enhance the zero-shot capabilities of SAM. However, they still suffer from limitations in tasks such as shadow detection, camouflaged detection, and medical imaging, where the boundaries between the foreground and background are unclear, leading to relatively poor zero-shot performance. In this context, we conclude that the most effective approach to transfer the knowledge of Enhanced ADL within SAM is by fine-tuning SAM directly using the initial mask generated by the Enhanced ADL .

\subsection{Fine-tuning SAM for Downstream Task}

SAM consists of an image encoder that embeds input images; a prompt encoder that embeds various types of prompts, such as masks, points, and bounding boxes; and a lightweight mask decoder that combines the encoded information to generate masks. Each module has tunable parameters. A straightforward method to fine-tune SAM is the full fine-tuning approach, which involves training all parameters. However, this requires training an enormous number of parameters and may lead to inadequate performance when the available data is scarce \cite{kong2023peeling}.
To address these issues, parameter efficient fine-tuning (PEFT) methods have been proposed \cite{chen2023sam, zhong2024convolution}. These approaches freeze the image encoder parameters while adding adapters within ViT blocks or incorporating parallel LoRA modules into the image encoder, training only a small number of parameters. 

However, despite reducing parameters, they require loading the entire model and using the image encoder’s values during both forward and backward passes because the modules are applied within the encoder. As such, the actual GPU memory usage and training time were not substantially reduced \cite{gu2024build}.
Otherwise, simple approach to SAM fine-tuning is to freeze parameters of some modules while training specific modules. Fine-tuning a mask decoder was demonstrated to be a simple yet highly effective method in the medical domain in \cite{yii2023data}. Accordingly, we adopt a fine-tuning approach in which only the lightweight mask decoder is fine-tuned while the remaining modules are frozen. This enables the proposed approach to be efficient, utilizing approximately 12 GB of VRAM with a ViT-B model and a batch size of 4. As a result, it functions effectively even when hardware resources are limited.

\section{Method}
\label{sec3}

\subsection{Overview}
\begin{figure*}[t] 
\centering
\includegraphics[width=1\linewidth]{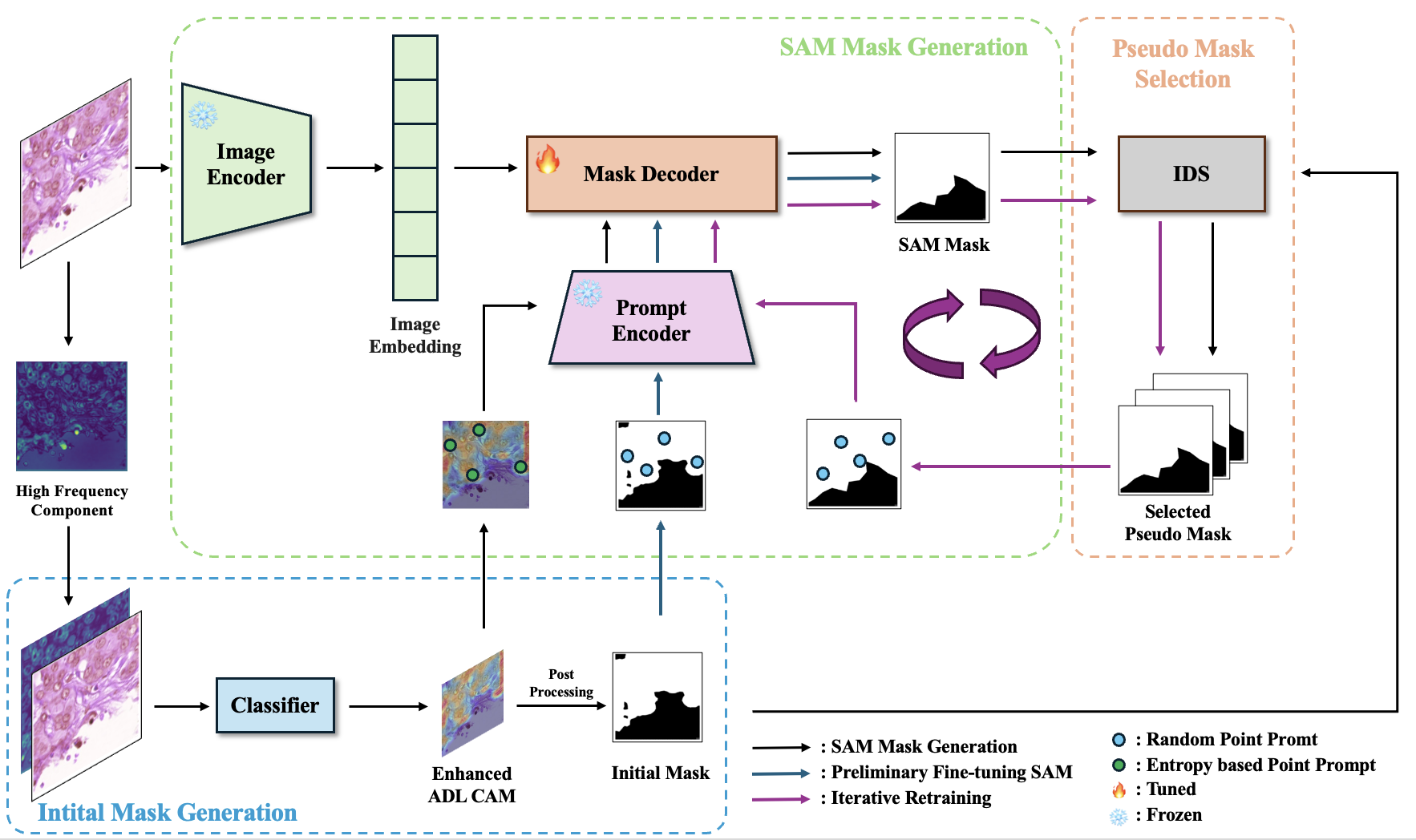}
\vspace{-1.5em}
\caption{Overview of our proposed method.}\label{fig1}
\end{figure*}

As depicted in Figure \ref{fig1}, the proposed method comprises three phases. First, an Enhanced ADL CAM is obtained from the patch classifier, and an initial mask is generated using a post-processing module. Second, the SAM mask decoder is fine-tuned using the initial mask, and a SAM pseudo-label is generated via a pixel-level entropy-based prompting module.  This also includes a filtering module that selects reliable pseudo-labels by assessing the intersection ratio between the SAM masks and initial masks. Finally, the selected pseudo-labels are used to fine-tune the re-initialized SAM mask decoder in iterative fashion.

\begin{figure}[t] 
\centering
\includegraphics[width=1\linewidth]{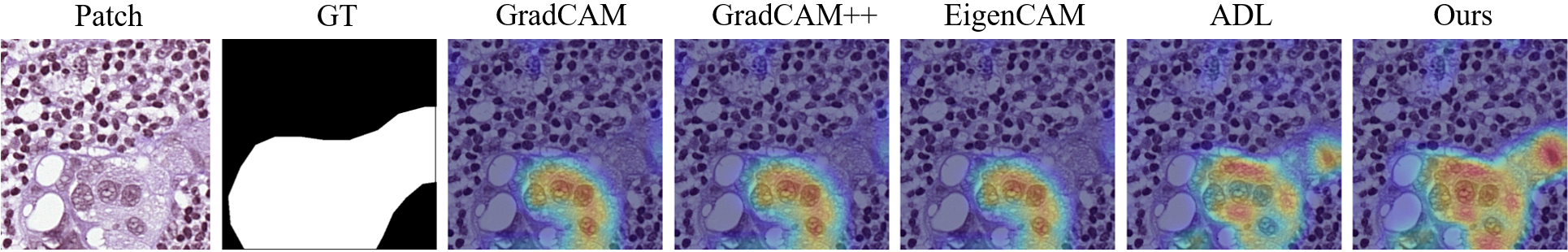}
\vspace{-1.5em}
\caption{Various CAMs for Camelyon17.}\label{fig2}
\end{figure}

\subsection{Initial Mask Generation Phase}

\subsubsection{Enhanced Attention Dropout Layer}
Generally, CAM tend to focus on the most discriminative part of an object rather than the entire object. On the other hand, ADL emphasizes broader regions by thresholding the attention map obtained by channel-wise pooling feature maps \( F \in \mathbb{R}^{C \times H \times W} \).

\begin{equation}
\label{equation2}
\begin{aligned}
M_{drop} &= 
\begin{cases} 
0 & \text{if } M_{att_{ij}} > \text{threshold} \\
1 & \text{otherwise}
\end{cases},
\quad
M_{imp} &= \sigma(M_{att}).
\end{aligned}
\end{equation}

The attention map $M_{att}$ is represented by \( M_{att} \in \mathbb{R}^{H \times W} \). Attention map produces either a drop mask or an importance map. The drop mask hides the most discriminative regions via thresholding, whereas the importance map highlights informative regions. Each drop mask $M_{drop}$ and importance map $M_{imp}$ is calculated using Eq. \eqref{equation2}.

Applying ADL to low-level feature maps like layer 1 and 2 reduces accuracy due to their unrelated to the target
\cite{choe2019attention}. Here, ADL is applied at layer 3’s first bottleneck and layer 4’s first bottlenecks in ResNet.

\begin{equation}
\label{equation3}
E_i = ADL(P_i, EVP_i). 
\end{equation}

Then, an explicit visual prompt is added to the patch image, which is subsequently used in ADL CAM, resulting in Enhanced ADL. In Eq. \eqref{equation3}, $E_{i}$ denotes the Enhanced ADL, $P_{i}$ represents the patches, and $EVP_{i}$  represents the explicit visual prompts that correspond to the high-frequency components extracted from the input data.

The Enhanced ADL CAM obtained in this way, as you can see in Figure \ref{fig2}, enables the classifier to consider the high frequency channel during training. When the CAM is extracted, this leads to more uniform and higher activation not only across the overall area of the target but also particularly around the blurred boundaries.

\subsubsection{Post-processing}
We also introduce a post-processing module designed to refine more precise CAM mask. This module employs quantile-based thresholding where the bottom $n$ of activation values (excluding zeros) are set to zero, while the remaining values are converted to one. Additionally, we utilize the \textit{rotate and fuse} technique along with morphological operations to achieve more accurate initial masks.

\textbf{Rotate \& Fuse} For reliable initial masks $I$, we employed rotation, a technique commonly used in data augmentation. For each enhanced image $E_i$, and patch \( P_i \), four CAMs \( \{E^k_i\}_{k=1}^K \) are generated by rotating the input through angles of \( 0^\circ, 90^\circ, 180^\circ, \) and \( 270^\circ \), corresponding to \( k = 1, 2, 3, 4 \) respectively. The CAMs are inversely rotated to their original orientation and averaged for the final result as follows: where $I$ denotes the final result (initial mask) and $i$ represents each patch index. $E_i$ denotes the enhanced image and $K$ indicates the rotation index corresponding to the angles \( 0^\circ, 90^\circ, 180^\circ, \) and \( 270^\circ \).

\begin{equation}
\label{equation4}
I_i = \frac{1}{K} \sum_{k=1}^K {E^k_i}.  
\end{equation}

\textbf{Morphological operation} DenseCRF is commonly used as a post-processing algorithm; however, it is highly sensitive to hyperparameters, which makes the search for optimal values inefficient. In addition, identifying a universally optimal value for an entire dataset is particularly challenging for medical images with unclear boundaries \cite{kervadec2020bounding}.

To resolve this, we apply a simple, effective post-processing technique using morphological operations, specifically opening, to remove small-scale noise. This operation comprises erosion and dilation using a structuring element. The erosion step removes small objects such as noises, and the subsequent dilation step restores the size of larger objects while avoiding the reappearance of small noises. An opening operation is employed to eliminate the noise generated by the threshold CAM, thereby yielding a more precise initial mask. The optimal configuration is then determined through a series of experiments.

\subsection{SAM Mask Generation Phase}
In this phase, we preliminarily train the SAM's decoder using the precise CAM mask obtained earlier, and to leverage the advantages of SAM, which can utilize prompts, we design a pixel-level entropy-based point prompting module using the Enhanced ADL. 

\subsubsection{Pixel-level Entropy based Prompting Module (PEPM)}
As demonstrated in \cite{cheng2023sam, gaus2024performance}, box prompts yield better results than other prompts for SAM; however, the box-prompt approach has several limitations. The conversion of CAM into discrete bounding boxes is sensitive to threshold configurations and requires extensive tuning to achieve optimal results. In addition, in cases where cancer regions are sparsely distributed within a pathology image, the 'best' bounding box prompt becomes a 'whole box' prompt, which fails to provide a helpful prompt for SAM in practice.

In contrast, \cite{cheng2023sam2d} demonstrated that SAM can achieve good performance even with multiple point prompts instead of bounding boxes, and \cite{dai2023samaug} showed that utilizing high entropy point prompts can enhance segmentation performance. In line with this, we have designed a pixel-level entropy-based point prompting module that leverages SAM's performance by taking advantage of the characteristic of Enhanced ADL, which provides uniform and high activation across the entire target area. If $A_{ij}$ denotes the activation obtained from the Enhanced ADL at a pixel, the entropy $S_{ij}$ of the pixel can be expressed as follows:
\begin{equation}
\label{equation5}
S_{ij} = \frac{A_{ij}}{\sum_{i=1}^n \sum_{j=1}^n A_{ij}}.
\end{equation}

\subsubsection{Preliminary SAM Mask Decoder Fine-tuning}
In medical images, Utilizing SAM in a zero-shot manner to generate pseudo labels significantly degrades the quality of the pseudo labels. \cite{mazurowski2023segment}. To address this issue, the SAM mask decoder is preliminarily trained using initial masks to effectively transfer knowledge from the Enhanced ADL. During this training, the SAM image encoder is frozen, and only the SAM mask decoder is fine-tuned.
\begin{figure}[t] 
\centering
\includegraphics[width=1\linewidth]{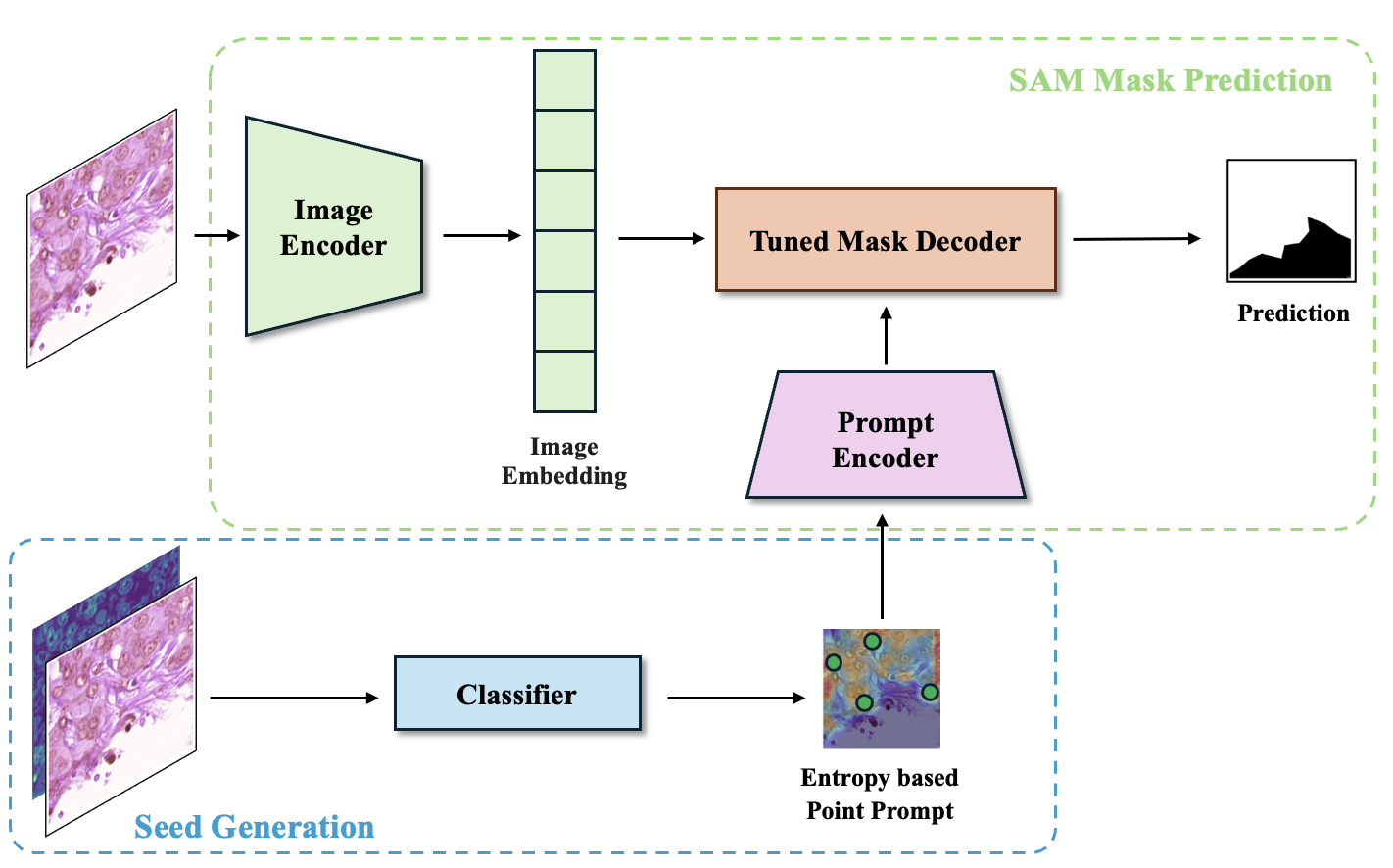}
\vspace{-1.5em}
\caption{Overall scenario in inference phase.}\label{fig3}
\end{figure}

\subsubsection{Pseudo-label Selection Module}
The pseudo-labels generated by SAM often contain noise, making it challenging to effectively generalize to medical images \cite{hu2023efficiently}. Thus, a more suitable approach for selecting appropriate pseudo-labels is required. Recent studies have utilized SAM for pseudo-labeling, and research has been conducted to produce high-quality pseudo-labels \cite{chen2023segment, wang2023cs, 
yang2024weakly}. These studies achieved promising results by employing an intersection ratio to address the incompleteness and redundancy inherent in initial CAM masks. Therefore, we utilized the intersection of the SAM mask and CAM mask divided by the SAM mask (IDS). Based on empirical comparisons, we set the threshold to 0.9.

\subsection{Prediction of Masks using Fine-tuned SAM}
Instead of using an off-the-shelf segmenter, the proposed method leverages SAM pseudo-labels to fine-tune SAM, which is then utilized as a mask predictor. This approach eliminates the need for additional training and maximizes the capability of SAM during inference by utilizing PEPM. The detailed inference procedure is illustrated in Figure \ref{fig3}.
\begin{figure}[t] 
\centering
\includegraphics[width=1\linewidth]{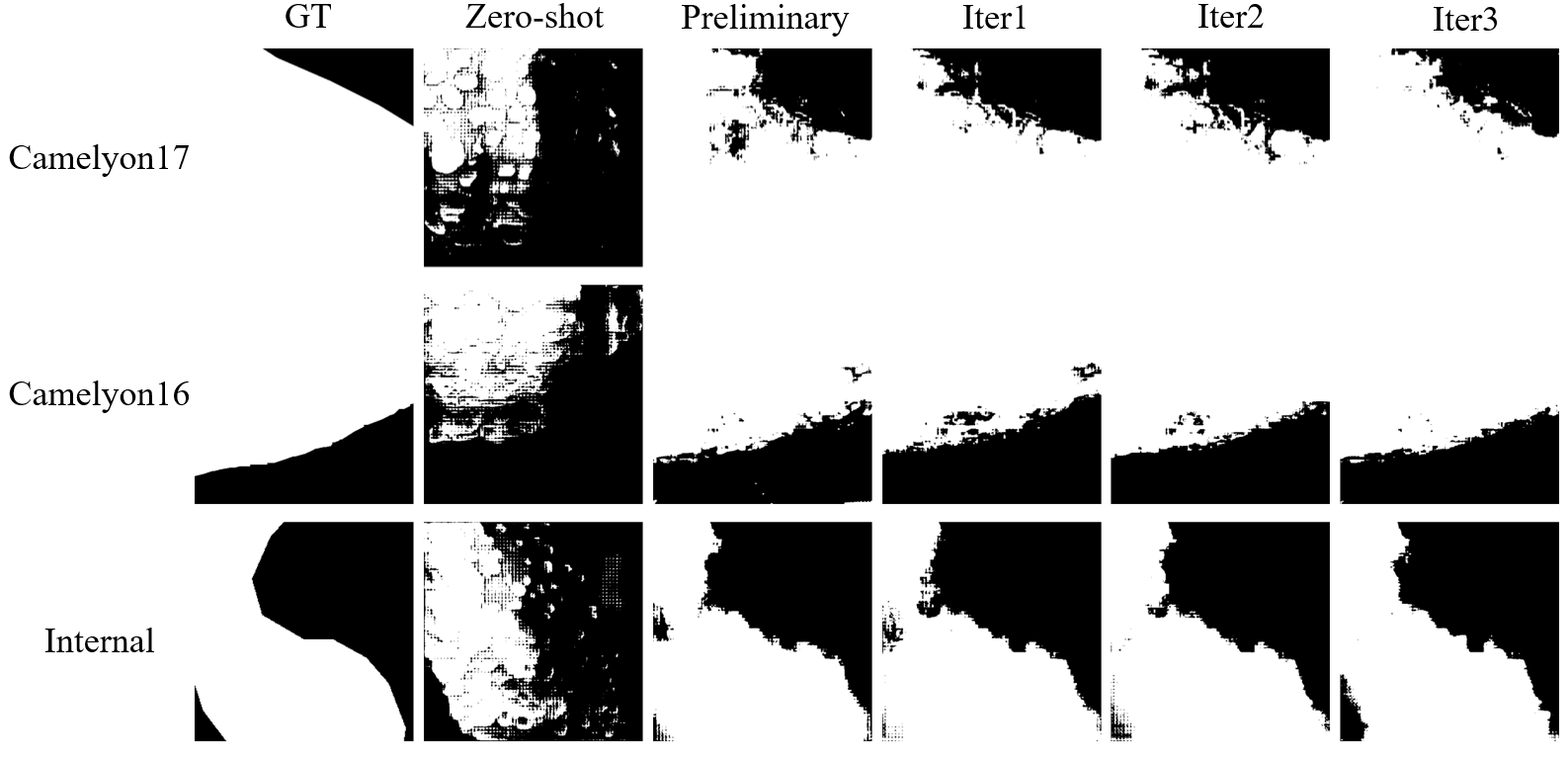}
\vspace{-1.5em}
\caption{Zero-shot results of SAM with respect to the iterative changes in masks. Progressive refinement is observed as training progresses, with the white tumor region becoming closer to the ground truth as iterations increase.}\label{fig4}
\end{figure}

\begin{algorithm}
\caption{Iterative Re-Training Strategy}
\label{al1}
\begin{algorithmic}
\Statex \textbf{Input:} Initial masks $\{I_k\}_{k=1}^K$, Threshold $t$, Number of iterations $N$
\Statex \textbf{Output:} Selected pseudo labels $P_L$
\Statex \hrule
    \Procedure{IterativeRetraining}{$\{I_k\}_{k=1}^K, t, N$}
        \State ${P_L} \gets \{\}$ \Comment{Saves selected pseudo labels}
        \For{$n = 1$ to $N$}
            \State $\{S_k\}_{k=1}^K \gets \text{SAM}(\{I_k\}_{k=1}^K)$ \Comment{Generate SAM masks}
            
            \For{$k = 1$ to $K$} 
                \State $IDS = \frac{\text{Intersect}(I_k, S_k)}{\text{nonzero\_area}(S_k)}$ 
                \If{$IDS > t$} 
                    \State $P_L \gets P_L \cup S_k$ 
                \EndIf
            \EndFor
            \State $\text{SAM.decoder\_init}()$ \Comment{Initialize SAM decoder}
            \State $W_n \gets \text{TrainDecoder}(P_L)$ \Comment{Train SAM decoder}
            \State $\text{SAM.decoder} \gets W_n$ \Comment{Update decoder weights}
        \EndFor
        \State \Return $P_L$ \Comment{Return selected pseudo labels}
    \EndProcedure
\end{algorithmic}
\end{algorithm}

\subsection{Iterative Retraining Phase}

As depicted in Figure \ref{fig4}, following preliminary fine-tuning, SAM is observed to generate close to the ground truth pseudo-labels by leveraging the knowledge transferred from the Enhanced ADL. Based on this observation, we hypothesize that retraining SAM iteratively using the enhanced SAM masks obtained via preliminary fine-tuning can further optimize its performance. The details of the retraining process are outlined in Algorithm \ref{al1}.
 
First, the initial mask $I_k$, whose generation is described in Section 3.2, is used as an input to SAM to produce the SAM mask. Then, high-quality pseudo-labels are obtained by filtering SAM masks that exceed the threshold $t$ using $IDS$, and selecting a pseudo-label $P_L$. Subsequently, the SAM mask decoder is trained using $P_L$, and trained SAM mask decoder weights $W_n$ are utilized to generate an enhanced pseudo-label $P_L$. The SAM mask decoder is initialized before each training session, and this process is iterated. As the iterations progress, the SAM mask decoder generates more robust pseudo-labels than the zero-shot SAM mask decoder.
\begin{figure*}[t] 
\centering
\includegraphics[width=1\linewidth]{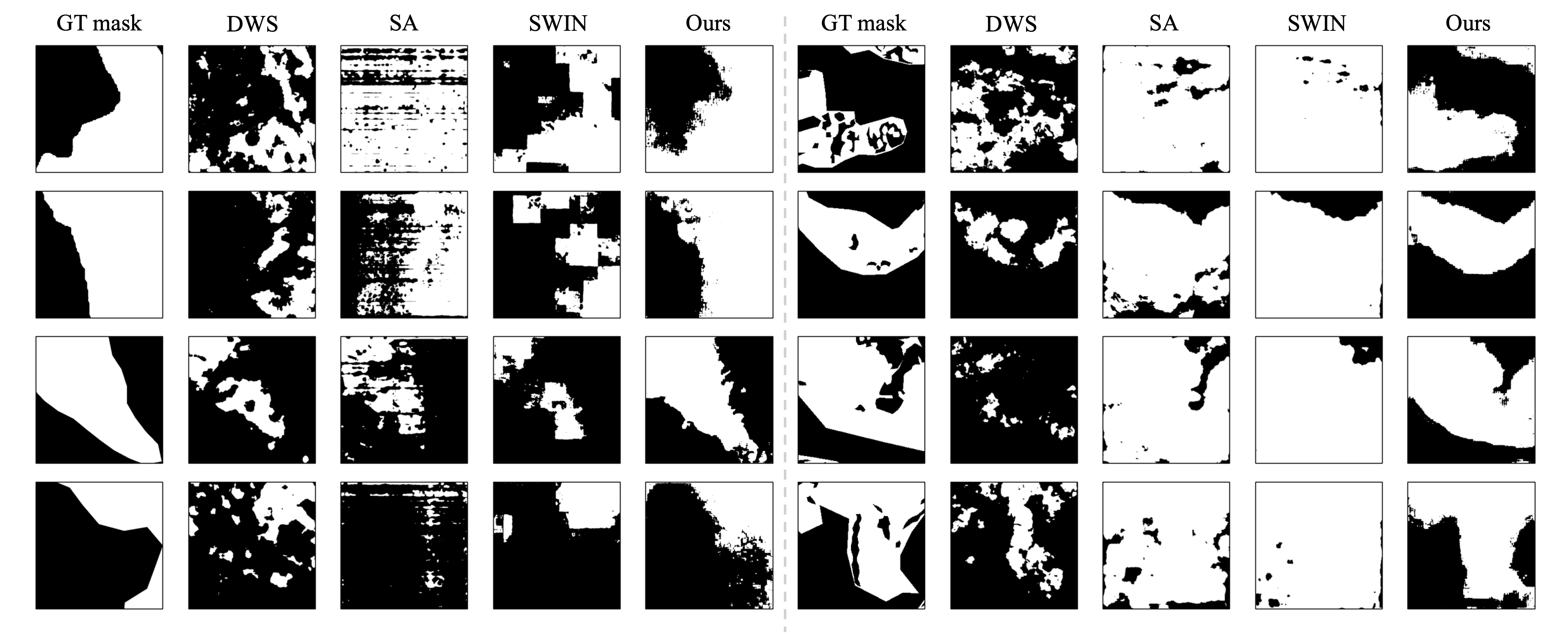}
\vspace{-1.5em}
\caption{Qualitative comparison between our proposed method and MIL-based methods across all datasets. Left: Camelyon16, Camelyon17 datasets. Right: internal dataset.}\label{fig9}
\end{figure*}
\section{Experimental Results and Discussion}
\label{sec4}

\subsection{Datasets}
Experiments were conducted to validate the proposed method on three histopathological breast cancer datasets. Two of the datasets used are open datasets—Camelyon 16 and Camelyon 17—and the third is an internal dataset.

The Camelyon16 dataset, provided by the Camelyon16 Challenge, is an open collection sourced from the Radboud University Medical Center and Utrecht University Medical Center. The slides are stained with hematoxylin and eosin (H\&E). The training dataset comprises 160 normal slides and 110 whole slide images (WSIs) depicting metastases. The test dataset comprises 130 WSIs. All slides are scanned at a magnification level of 40x to providing high-resolution images for detailed analysis.

The Camelyon17 dataset, provided by the Camelyon17 Challenge, the successor to Camelyon16, is collected from five centers: Radboud University Medical Center, Utrecht University Medical Center, Rijnstate Hospital, Canisius-Wilhelmina Hospital, and LabPON. It is significantly larger than the Camelyon16 dataset, and offers a comprehensive collection of 1000 WSIs. This extensive dataset enhances the potential for robust training and validation of machine-learning models designed for histopathological analysis.

For Camelyon16 and Camelyon17, the dataset was constructed by extracting patches from WSIs that were positive only. The patch size was uniformly set to 512 × 512 pixels, and both positive and negative patches were extracted using a sliding window with a stride of 256 pixels. From the candidate pool of positive patches, only those in which the tumor occupies  20\% and 90\% of the area were selected. From the candidate pool of positive patches, only those in which the tumor occupies 20–90\% of the area were selected.

Table \ref{table8} lists the number of positive and negative patches in each dataset. To balance each dataset, an equivalent number of negative patches are selected to match the number of positive patches. The data leakage was prevented by ensuring that the WSIs in the train, validation, and test sets did not overlap.
\begin{table}[htbp]
\centering
\resizebox{0.4\textwidth}{!}{%
\begin{tabular}{@{}ccccccc@{}}
\toprule
Dataset & \multicolumn{2}{c}{Camelyon16} & \multicolumn{2}{c}{Camelyon17} & \multicolumn{2}{c}{Internal} \\
\cmidrule(lr){2-3} \cmidrule(lr){4-5} \cmidrule(l){6-7}
Data splits & \textit{Pos} & \textit{Neg} & \textit{Pos} & \textit{Neg} & \textit{Pos} & \textit{Neg} \\
\midrule
Train       & 6020         & 6000         & 6068         & 6000         & 3111         & 3111         \\
Valid       & 700          & 700          & 698          & 700          & 120          & 120          \\
Test        & 2002         & 2000         & 2000         & 2000         & 350          & 350          \\
\bottomrule
\end{tabular}
}
\caption{Number of positive and negative patches in each dataset split.}
\label{table8}
\end{table}
\subsection{Implementation Details}
In our implementation, the SAM image encoder is based on \textit{ViT-B/16}. The backbone of the patch classifier is taken to be \textit{ResNet50}, which is also employed to train other classifier for CAM extraction during comparative evaluation. The classifier training employs Binary Cross Entropy loss, using the Adam optimizer, a learning rate of $1 \times 10^{-5}$, weight decay of $1 \times 10^{-3}$, a batch size of 16. Training is conducted over 50 epochs. To fine-tune the SAM mask decoder, a linear combination of Dice loss and intersection-over-union (IoU) loss is used as the loss function. AdamW is used as the optimizer, with a learning rate of $2 \times 10^{-4}$, and the model is trained for 20 epochs.

All experiments are performed using PyTorch on single NVIDIA TITAN Xp. Additionally, for comparative experiments, single NVIDIA A6000 is used.
\begin{table*}[t] 
\centering
\begin{tabular}{ccccccccc}
\toprule
\multirow{2}{*}{Model} & \multirow{2}{*}{Backbone} & \multirow{2}{*}{SUP} & \multicolumn{2}{c|}{Camelyon17} & \multicolumn{2}{c|}{Camelyon16} & \multicolumn{2}{c}{Internel} \\ \cmidrule(l){4-9} 
                       &                           &                      & Dice (\%)            & \multicolumn{1}{c|}{IoU (\%)} & Dice (\%)            & \multicolumn{1}{c|}{IoU (\%)} & Dice (\%)            & IoU (\%)             \\ \midrule
\multicolumn{9}{c}{Full Supervision}                                                                                                                                  \\ \midrule
U-Net \cite{ronneberger2015u}                  & ResNet50                  &  $F$                   & 82                   & \multicolumn{1}{c|}{72.88}    & 73.21                & \multicolumn{1}{c|}{61.05}    & 86.28       & 76.76       \\
SAM-Decoder (Whole box) \cite{kirillov2023segment} & ViT-B                    & $F$                     & 83.72       & \multicolumn{1}{c|}{73.95} & 78.26       & \multicolumn{1}{c|}{66.56} & 81.28                & 69.74                \\
MedSAM-Decoder (Whole box) \cite{ma2024segment} & ViT-B                &  $F$                    & 81.06                & \multicolumn{1}{c|}{70.07}    & 68.69                & \multicolumn{1}{c|}{54.99}    & 81.08                & 70.07                \\ \midrule
\multicolumn{9}{c}{Weak Supervision}                                                                                                                                \\ \midrule
CAM based              &                           &                      &                      &                             &                      &                             &                      &                      \\ \midrule
GradCAM \cite{selvaraju2020grad}              & ResNet50                  &  $I$                    & 56.48                & \multicolumn{1}{c|}{40.24}    & 63.36                & \multicolumn{1}{c|}{47.76}    & 70.13                & 55.09                \\
GradCAM++ \cite{chattopadhay2018grad}             & ResNet50                  & $I$                      & 63.16                & \multicolumn{1}{c|}{47.06}    & 65.82                & \multicolumn{1}{c|}{50.22}    & 70.54                & 55.55                \\
EigenCAM \cite{muhammad2020eigen}              & ResNet50                  & $I$                    & 56.09                & \multicolumn{1}{c|}{40.04}    & 59.4                 & \multicolumn{1}{c|}{43.4}     & 66.72                & 51.63                \\
ADL \cite{choe2019attention}                   & ResNet50                  &   $I$                    & 79.39                & \multicolumn{1}{c|}{67.31}    & 70.6        & \multicolumn{1}{c|}{56.89} & 69.79                & 54.37                \\
Enhanced ADL           & ResNet50                  &   $I$                    & 80.1        & \multicolumn{1}{c|}{68.21} & 69.61                & \multicolumn{1}{c|}{55.59}    & 72.76       & 58.18       \\ \midrule
WSSS Methods           &                           &                      &                      &                             &                      &                             &                      &                      \\ \midrule
U-Net \cite{ronneberger2015u}                 & ResNet50                     & $P$                  & 79.91                   & \multicolumn{1}{c|}{69.16}      & 75.85                   & \multicolumn{1}{c|}{62.83}      & 72.91                   & 59.87                   \\
WSSS-Tissue \cite{han2022multi}           & ResNet38-D                  &   $I$                   & 33.16                & \multicolumn{1}{c|}{20.88}    & 46.92                & \multicolumn{1}{c|}{32.24}    & 68.85                & 54.35                \\
Swin-MIL \cite{qian2022transformer}              & VGG16                     &  $I$                     & 66.6                 & \multicolumn{1}{c|}{60.9}     & 54.9                 & \multicolumn{1}{c|}{48.6}     & 55.4                 & 48.8                 \\
DWS-MIL \cite{jia2017constrained}                & VGG16                     & $I$                     & 39.3                 & \multicolumn{1}{c|}{32.2}     & 32                   & \multicolumn{1}{c|}{21.9}     & 38.7                 & 32                   \\
SA-MIL \cite{li2023weakly}                & VGG16                     & $I$                       & 58.9                 & \multicolumn{1}{c|}{52.5}     & 58.7                 & \multicolumn{1}{c|}{52.2}     & 57.1                 & 50.8                 \\
Ours                   & ViT-B                     &  $I$                     & \textbf{83.83}       & \multicolumn{1}{c|}{\textbf{73.74}} & \textbf{76.94}       & \multicolumn{1}{c|}{\textbf{64.99}} & \textbf{75.13}       & \textbf{61.5}        \\ \bottomrule
\end{tabular}
\caption{Performance comparison across different models, backbones, and supervision levels on three datasets. The SUP. column indicates the form of supervision applied during training, encompassing full supervision ($F$), training with pseudo labels ($P$), and image-level labels ($I$).}
\label{table1}
\vspace{-1em} 
\end{table*}
\subsection{Comparative Evaluation}

We conducted evaluations using open datasets, including Camelyon16, Camelyon17, as well as an internal dataset. We compared the proposed method and existing MIL-based SOTA methods as well as fully supervised methods. Moreover, given the high capacity of SAM, we compared Unet and MedSAM as fully supervised manner, which are widely used in the medical domain. For MedSAM training, only the mask decoder was fine-tuned, while all other settings followed MedSAM's original configuration. In addition, various CAM-based methods are evaluated. To ensure a fair comparison, the post-processing used to our framework is also applied to the CAM variants,  and the hyperparameters are optimized via a grid search, with the best values reported. Further, the generalization performance of the pseudo-labels generated by our framework is evaluated by training a U-Net segmenter.

As highlighted in Table \ref{table1}, the proposed method significantly outperforms existing MIL-based SOTA and CAM-based methods on all datasets. Notably, on the Camelyon17 and Camelyon16 datasets, the proposed approach also outperforms the fully supervised models, MedSAM and Unet. Further, when the pseudo-labels generated by our framework are used for training, the performance gap compared with fully supervised learning method is observed to be less than  3\%.
\begin{figure}[t] 
\centering
\includegraphics[width=1\linewidth]{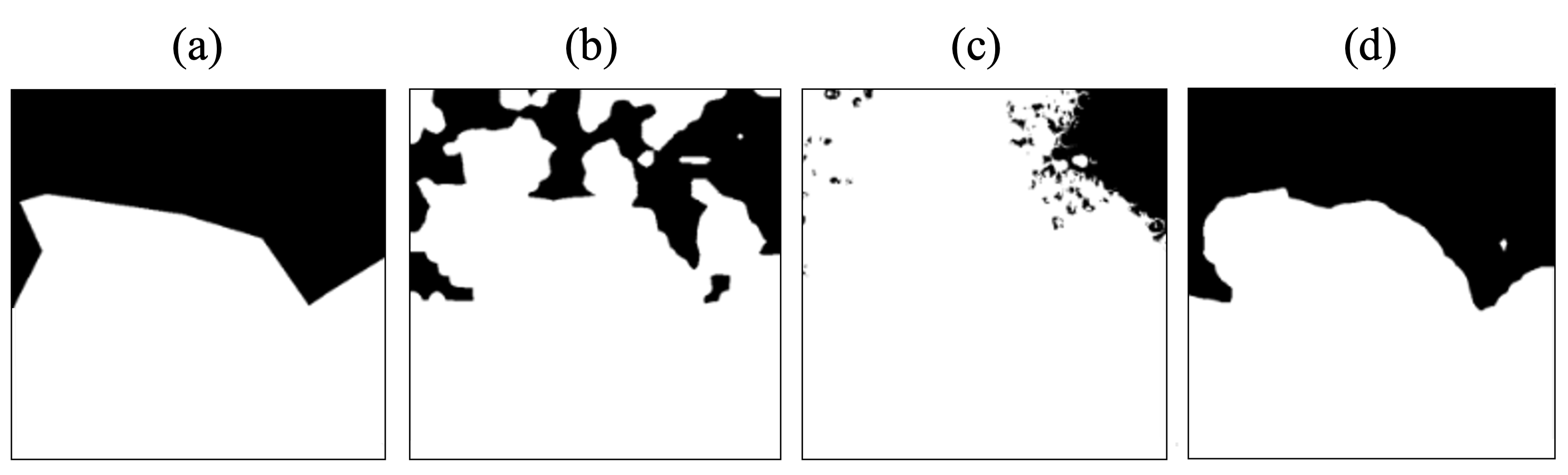}
\vspace{-1.5em}
\caption {Comparative results of different processing techniques on the Camelyon17 dataset. (a) displays the ground truth (GT), (b) shows results from the Enhanced ADL without post-processing, (c) depicts the outcomes of applying DenseCRF to the Enhanced ADL and (d) illustrates results from our method.}\label{fig6}
\end{figure}
Figure \ref{fig9} illustrates a comparison of outputs between the MIL-based SOTA method used in our comparative experiments and our proposed method across each dataset. Through comparison with the GT mask, we can confirm that our proposed method demonstrates superior performance relative to other methods. Even in the case of Camelyon16 and Camelyon17, which present relatively lower segmentation difficulty, we can observe that the output results of the MIL-based method differ significantly when compared to the GT mask. In some cases, the overall shape of the GT mask is identified with reasonable similarity. Howerver, we can also observe instances where methods like SA-MIL produce entirely incorrect results. Furthermore, we can identify specific unintended patterns in the output results, such as horizontal lines in SA-MIL and rectangles in SWIN-MIL. We can also observe cases like DWS-MIL where only partial regions are detected, failing to identify the entire area of interest. In contrast, our proposed method demonstrates output results that are generally similar to the GT mask, with the exception of some inaccuracies at the boundaries. For the internal dataset, SA-MIL and SWIN-MIL incorrectly classified almost the entire input patch as positive class. While DWS-MIL identified a similar overall shape of the positive regions, it demonstrated difficulties in detecting the entire region of interest or misclassified negative areas as positive class, similar to its performance on the Camelyon dataset. Our method, in contrast, demonstrates accurate identification of positive regions. Although it may not precisely capture small, detailed areas within the tumor, it shows excellent results when compared to other MIL-based methods.

\subsection{Ablation Study}

\renewcommand{\thesubsubsection}{\Alph{subsubsection}}
\subsubsection{Effectiveness of Mask Generation Module}

We conducted an ablation study to evaluate the effectiveness of our post-processing technique. Additionally, we included denseCRF, one of the most commonly used post-processing methods, for comparison. As shown in Table \ref{table2}, we found that incorporating all components resulted in the best performance across both datasets. Furthermore, denseCRF demonstrated lower performance compared to ADL with post-processing and even Enhanced ADL without post-processing. Our proposed post-processing technique demonstrated superior denoising capacity compared to denseCRF, as further illustrated by the visual comparisons in Figure \ref{fig6}.

\begin{figure}[t] 
\centering
\includegraphics[width=1\linewidth]{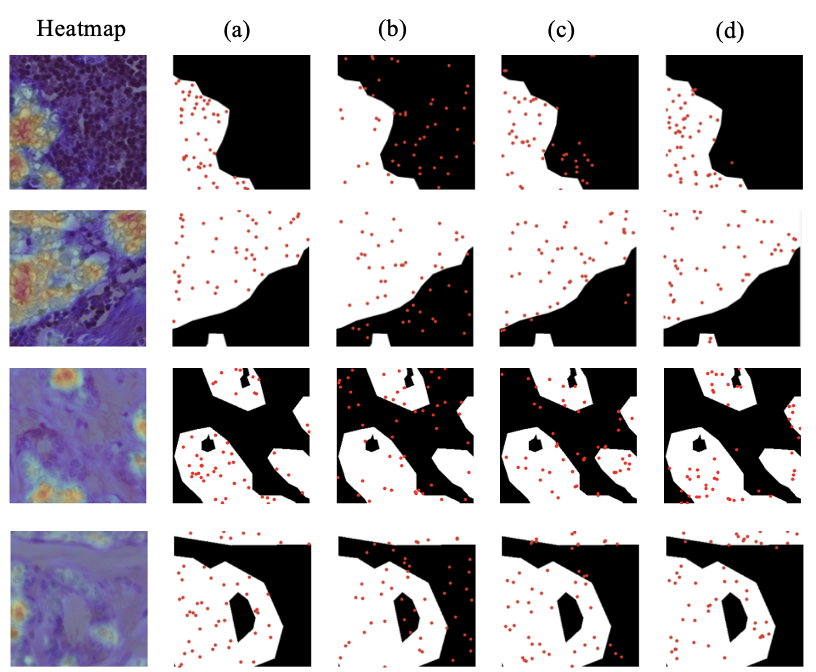}
\vspace{-1.5em}
\caption{Qualitative comparison of the effects of PEPM. The bottom two rows correspond to the internal dataset, and the top two rows to the Camelyon17 dataset. (a) Fifty random points from the ground truth; (b) fifty random points from the entire area; (c) fifity random points from CAM-activated areas;, and (d) fifty points based on the proposed PEPM method.}\label{fig5}
\end{figure}

\begin{table}[htbp]
\centering
\resizebox{0.5\textwidth}{!}{%
\begin{tabular}{c|cc|cc|cc}
\hline
\multirow{2}{*}{Method}        & \multicolumn{2}{c|}{Post-Processing} & \multicolumn{2}{c|}{Camelyon17} & \multicolumn{2}{c}{Internel}    \\ \cline{2-7} 
                              & Ours              & CRF              & Dice (\%)      & IoU (\%)       & Dice (\%)      & IoU (\%)       \\ \hline
ADL                           & $\checkmark$                  &                  & 78.11          & 65.70          & 72.57          & 57.60          \\ \hline
\multirow{3}{*}{Enhanced ADL} &                   &                  & 70.12          & 56.68          & 68.15          & 53.53          \\
                              &                   & $\checkmark$                 & 69.63          & 56.16          & 67.72          & 53.15          \\
                              & $\checkmark$                  &                  & \textbf{78.67} & \textbf{66.42} & \textbf{76.33} & \textbf{62.69} \\ \hline
\end{tabular}
}
\caption{The results of the ablation studies on the post-processing modules.}
\label{table2}
\end{table}

\begin{table}[htbp]
\centering
\small 
\resizebox{0.3\textwidth}{!}{%
\begin{tabular}{@{}c|cc@{}}
\toprule
Method            & Dice (\%) & IoU (\%) \\ \midrule
GT Random points  & 81.01     & 69.07    \\
Random points     & 69.24     & 54.86    \\
w/o PEPM & 71.75     & 57.65    \\
Ours              & 76.57     & 63.16    \\ \bottomrule
\end{tabular}
}
\caption{The results of the ablation studies to assess the effectiveness of PEPM on internal data.}
\label{table5}
\end{table}
\begin{table}[htbp]
\centering
\resizebox{0.5\textwidth}{!}{%
\begin{tabular}{c|cc|cc|cc}
\hline
\multirow{2}{*}{Iter} & \multicolumn{2}{c|}{Camelyon17} & \multicolumn{2}{c|}{Camelyon16} & \multicolumn{2}{c}{Internel} \\ \cline{2-7} 
                      & Dice (\%)       & IoU (\%)      & Dice (\%)       & IoU (\%)      & Dice (\%)     & IoU (\%)     \\ \hline
Zero-shot             & 34.24           & 22.63         & 35.51           & 23.92         & 43.89         & 29.34        \\
preliminary           & 78.66           & 66.41         & 70.38           & 56.55         & 75.06         & 61.16        \\
1                     & 82.16           & 71.26         & 75.20           & 62.63         & 76.57         & 63.16        \\
2                     & 82.95           & 72.40         & 75.83           & 63.77         & 77.03         & 63.83        \\
3                     & \textbf{82.98}  & \textbf{72.42} & \textbf{76.41} & \textbf{64.47} & \textbf{77.47} & \textbf{64.40} \\ \hline
\end{tabular}
}
\caption{Quantitatively evaluated the quality of the pseudo labels generated at each iteration across the Camelyon17, Camelyon16, and internal datasets.}
\label{table3}
\end{table}

\subsubsection{Effectiveness of PEPM}
We also conducted an ablation study to evaluate the effectiveness of the PEPM module. As shown in Table \ref{table5}, using the PEPM module outperformed other methods, such as randomly selecting points from the entire input patch or generating random points as prompts after post-processing without the PEPM module. 
This indicates that our proposed PEPM, by providing points around the boundary area as prompts, enables SAM to effectively depict blurred boundaries in histopathology images.

Furthermore, Figure \ref{fig5} illustrates that our module generates better seeds based on entropy, effectively capturing boundaries and confirming its superior performance.

\begin{table}[htbp]
\centering
\resizebox{0.5\textwidth}{!}{%
\begin{tabular}{@{}c|cc|cc|cc@{}}
\toprule
\multirow{2}{*}{Iter} & \multicolumn{2}{c|}{Camelyon17} & \multicolumn{2}{c|}{Camelyon16} & \multicolumn{2}{c}{Internel} \\ \cmidrule(l){2-7} 
                      & Dice (\%)       & IoU (\%)      & Dice (\%)       & IoU (\%)      & Dice (\%)     & IoU (\%)     \\ \midrule
Preliminary           & 83.19           & 72.7          & 75.65           & 63.15         & 76.97         & 63.69        \\
1                     & 83.65           & 73.32         & 76.44           & 64.40         & 77.68         & 64.70        \\
2                     & 84.01           & 73.79         & 76.84           & 64.82         & 78.30         & 65.47        \\
3                     & 84.15           & 74.09         & 76.81           & 64.73         & 78.57         & 65.85      
    \\ \bottomrule
\end{tabular}
}
\caption{Impact of retraining strategy we evaluate mIOU (\&) and mDice (\%) on the \textit{validation sets} of three datasets.}
\label{table4}
\end{table}
\subsubsection{Effectiveness of Retraining Module}
As shown in Table \ref{table3}, the preliminary fine-tuning results showed approximately twice the performance compared to the zero-shot outcomes. Furthermore, the quality of the pseudo labels continued to improve with subsequent iterations. These results demonstrated the effectiveness of our proposed preliminary fine-tuning strategy.
We also verified the effectiveness of the retraining strategy by quantitatively analyzing the quality of the pseudo labels and the predicted masks for each iteration. In Table \ref{table4}, the metrics consistently improved with each iteration, further validating the success of our retraining approach.
As observed in Tables \ref{table3} and \ref{table4}, the metrics consistently improve with each iteration, underscoring the overall effectiveness of our proposed approach. Effectiveness of the retraining module can also be observed in Figure \ref{fig4}.

\section{Conclusion}

In this paper, we propose a weakly supervised semantic segmentation framework to address the high-cost labeling problem commonly encountered in whole slide image (WSI) segmentation scenarios. We utilized a classifier trained solely on patch-level annotations to generate CAM (Class Activation Map) masks for each patch. These masks were then employed in training a Segment Anything for the creation of pseudo labels. To generate high-quality CAM masks, we enhanced Attention Dropout Layers by incorporating explicit visual promptings technique, and simple but effective post-processing modules. For the creation of high-quality pseudo labels, we utilized a pixel-level entropy based prompting module, preliminary mask decoder fine-tuning, and an iterative retraining strategy. Experimental results demonstrate that our proposed framework outperforms both CAM-based methods and MIL-based state-of-the-art methods across all datasets. In several instances, it even surpasses the performance of fully supervised models. Furthermore, an ablation study was conducted, which conclusively showed the effectiveness of the proposed modules. All proposed structures are executable within 12GB of GPU memory, allowing for efficient performance of all processes without the requirement of high-performance hardware. Consequently, this accessibility is expected to result in high applicability in real-world industrial settings.
\label{sec5}


\begin{thebibliography}{50}
\expandafter\ifx\csname natexlab\endcsname\relax\def\natexlab#1{#1}\fi
\providecommand{\url}[1]{\texttt{#1}}
\providecommand{\path}[1]{#1}
\providecommand{\DOIprefix}{doi:}
\providecommand{\ArXivprefix}{arXiv:}
\providecommand{\URLprefix}{URL: }
\providecommand{\Pubmedprefix}{pmid:}
\providecommand{\doi}[1]{\href{http://dx.doi.org/#1}{\path{#1}}}
\providecommand{\Pubmed}[1]{\href{pmid:#1}{\path{#1}}}
\providecommand{\bibinfo}[2]{#2}
\ifx\xfnm\relax \def\xfnm[#1]{\unskip,\space#1}\fi
\bibitem[{Siegel et~al.(2024)Siegel, Giaquinto, and Jemal}]{siegel2024cancer}
\bibinfo{author}{R.~L. Siegel}, \bibinfo{author}{A.~N. Giaquinto}, \bibinfo{author}{A.~Jemal},
\newblock \bibinfo{title}{Cancer statistics, 2024.},
\newblock \bibinfo{journal}{CA: a cancer journal for clinicians} \bibinfo{volume}{74} (\bibinfo{year}{2024}).
\bibitem[{Ronneberger et~al.(2015)Ronneberger, Fischer, and Brox}]{ronneberger2015u}
\bibinfo{author}{O.~Ronneberger}, \bibinfo{author}{P.~Fischer}, \bibinfo{author}{T.~Brox},
\newblock \bibinfo{title}{U-net: Convolutional networks for biomedical image segmentation},
\newblock in: \bibinfo{booktitle}{Medical image computing and computer-assisted intervention--MICCAI 2015: 18th international conference, Munich, Germany, October 5-9, 2015, proceedings, part III 18}, \bibinfo{organization}{Springer}, \bibinfo{year}{2015}, pp. \bibinfo{pages}{234--241}.
\bibitem[{Xie et~al.(2021)Xie, Wang, Yu, Anandkumar, Alvarez, and Luo}]{xie2021segformer}
\bibinfo{author}{E.~Xie}, \bibinfo{author}{W.~Wang}, \bibinfo{author}{Z.~Yu}, \bibinfo{author}{A.~Anandkumar}, \bibinfo{author}{J.~M. Alvarez}, \bibinfo{author}{P.~Luo},
\newblock \bibinfo{title}{Segformer: Simple and efficient design for semantic segmentation with transformers},
\newblock \bibinfo{journal}{Advances in neural information processing systems} \bibinfo{volume}{34} (\bibinfo{year}{2021}) \bibinfo{pages}{12077--12090}.
\bibitem[{Chen et~al.(2017{\natexlab{a}})Chen, Papandreou, Kokkinos, Murphy, and Yuille}]{chen2017deeplab}
\bibinfo{author}{L.-C. Chen}, \bibinfo{author}{G.~Papandreou}, \bibinfo{author}{I.~Kokkinos}, \bibinfo{author}{K.~Murphy}, \bibinfo{author}{A.~L. Yuille},
\newblock \bibinfo{title}{Deeplab: Semantic image segmentation with deep convolutional nets, atrous convolution, and fully connected crfs},
\newblock \bibinfo{journal}{IEEE transactions on pattern analysis and machine intelligence} \bibinfo{volume}{40} (\bibinfo{year}{2017}{\natexlab{a}}) \bibinfo{pages}{834--848}.
\bibitem[{Chen et~al.(2017{\natexlab{b}})Chen, Papandreou, Schroff, and Adam}]{chen2017rethinking}
\bibinfo{author}{L.-C. Chen}, \bibinfo{author}{G.~Papandreou}, \bibinfo{author}{F.~Schroff}, \bibinfo{author}{H.~Adam},
\newblock \bibinfo{title}{Rethinking atrous convolution for semantic image segmentation},
\newblock \bibinfo{journal}{arXiv preprint arXiv:1706.05587}  (\bibinfo{year}{2017}{\natexlab{b}}).
\bibitem[{Zhang et~al.(2022)Zhang, Tian, Tang, Chu, Wei, Shen et~al.}]{zhang2022segvit}
\bibinfo{author}{B.~Zhang}, \bibinfo{author}{Z.~Tian}, \bibinfo{author}{Q.~Tang}, \bibinfo{author}{X.~Chu}, \bibinfo{author}{X.~Wei}, \bibinfo{author}{C.~Shen}, et~al.,
\newblock \bibinfo{title}{Segvit: Semantic segmentation with plain vision transformers},
\newblock \bibinfo{journal}{Advances in Neural Information Processing Systems} \bibinfo{volume}{35} (\bibinfo{year}{2022}) \bibinfo{pages}{4971--4982}.
\bibitem[{Strudel et~al.(2021)Strudel, Garcia, Laptev, and Schmid}]{strudel2021segmenter}
\bibinfo{author}{R.~Strudel}, \bibinfo{author}{R.~Garcia}, \bibinfo{author}{I.~Laptev}, \bibinfo{author}{C.~Schmid},
\newblock \bibinfo{title}{Segmenter: Transformer for semantic segmentation},
\newblock in: \bibinfo{booktitle}{Proceedings of the IEEE/CVF international conference on computer vision}, \bibinfo{year}{2021}, pp. \bibinfo{pages}{7262--7272}.
\bibitem[{Lin et~al.(2023)Lin, Qu, Chen, Gao, Li, Xia, Ma, Zheng, and Cheng}]{lin2023nuclei}
\bibinfo{author}{Y.~Lin}, \bibinfo{author}{Z.~Qu}, \bibinfo{author}{H.~Chen}, \bibinfo{author}{Z.~Gao}, \bibinfo{author}{Y.~Li}, \bibinfo{author}{L.~Xia}, \bibinfo{author}{K.~Ma}, \bibinfo{author}{Y.~Zheng}, \bibinfo{author}{K.-T. Cheng},
\newblock \bibinfo{title}{Nuclei segmentation with point annotations from pathology images via self-supervised learning and co-training},
\newblock \bibinfo{journal}{Medical Image Analysis} \bibinfo{volume}{89} (\bibinfo{year}{2023}) \bibinfo{pages}{102933}.
\bibitem[{Srinidhi et~al.(2022)Srinidhi, Kim, Chen, and Martel}]{srinidhi2022self}
\bibinfo{author}{C.~L. Srinidhi}, \bibinfo{author}{S.~W. Kim}, \bibinfo{author}{F.-D. Chen}, \bibinfo{author}{A.~L. Martel},
\newblock \bibinfo{title}{Self-supervised driven consistency training for annotation efficient histopathology image analysis},
\newblock \bibinfo{journal}{Medical image analysis} \bibinfo{volume}{75} (\bibinfo{year}{2022}) \bibinfo{pages}{102256}.
\bibitem[{Pati et~al.(2023)Pati, Jaume, Ayadi, Thandiackal, Bozorgtabar, Gabrani, and Goksel}]{pati2023weakly}
\bibinfo{author}{P.~Pati}, \bibinfo{author}{G.~Jaume}, \bibinfo{author}{Z.~Ayadi}, \bibinfo{author}{K.~Thandiackal}, \bibinfo{author}{B.~Bozorgtabar}, \bibinfo{author}{M.~Gabrani}, \bibinfo{author}{O.~Goksel},
\newblock \bibinfo{title}{Weakly supervised joint whole-slide segmentation and classification in prostate cancer},
\newblock \bibinfo{journal}{Medical Image Analysis} \bibinfo{volume}{89} (\bibinfo{year}{2023}) \bibinfo{pages}{102915}.
\bibitem[{Liang-Chieh et~al.(2015)Liang-Chieh, Papandreou, Kokkinos, Murphy, and Yuille}]{liang2015semantic}
\bibinfo{author}{C.~Liang-Chieh}, \bibinfo{author}{G.~Papandreou}, \bibinfo{author}{I.~Kokkinos}, \bibinfo{author}{K.~Murphy}, \bibinfo{author}{A.~Yuille},
\newblock \bibinfo{title}{Semantic image segmentation with deep convolutional nets and fully connected crfs},
\newblock in: \bibinfo{booktitle}{International conference on learning representations}, \bibinfo{year}{2015}.
\bibitem[{Gao et~al.(2021)Gao, Wan, Pan, Peng, Tian, Han, Zhou, and Ye}]{gao2021ts}
\bibinfo{author}{W.~Gao}, \bibinfo{author}{F.~Wan}, \bibinfo{author}{X.~Pan}, \bibinfo{author}{Z.~Peng}, \bibinfo{author}{Q.~Tian}, \bibinfo{author}{Z.~Han}, \bibinfo{author}{B.~Zhou}, \bibinfo{author}{Q.~Ye},
\newblock \bibinfo{title}{Ts-cam: Token semantic coupled attention map for weakly supervised object localization},
\newblock in: \bibinfo{booktitle}{Proceedings of the IEEE/CVF international conference on computer vision}, \bibinfo{year}{2021}, pp. \bibinfo{pages}{2886--2895}.
\bibitem[{Gu et~al.(2024)Gu, Wu, Tang, Mai, Shu, Li, and Chen}]{gu2024lesam}
\bibinfo{author}{Y.~Gu}, \bibinfo{author}{Q.~Wu}, \bibinfo{author}{H.~Tang}, \bibinfo{author}{X.~Mai}, \bibinfo{author}{H.~Shu}, \bibinfo{author}{B.~Li}, \bibinfo{author}{Y.~Chen},
\newblock \bibinfo{title}{Lesam: Adapt segment anything model for medical lesion segmentation},
\newblock \bibinfo{journal}{IEEE Journal of Biomedical and Health Informatics}  (\bibinfo{year}{2024}).
\bibitem[{Fang et~al.(2023)Fang, Chen, Wang, Wang, Ji, and Zhang}]{fang2023weakly}
\bibinfo{author}{Z.~Fang}, \bibinfo{author}{Y.~Chen}, \bibinfo{author}{Y.~Wang}, \bibinfo{author}{Z.~Wang}, \bibinfo{author}{X.~Ji}, \bibinfo{author}{Y.~Zhang},
\newblock \bibinfo{title}{Weakly-supervised semantic segmentation for histopathology images based on dataset synthesis and feature consistency constraint},
\newblock in: \bibinfo{booktitle}{Proceedings of the AAAI Conference on Artificial Intelligence}, volume~\bibinfo{volume}{37}, \bibinfo{year}{2023}, pp. \bibinfo{pages}{606--613}.
\bibitem[{Mazurowski et~al.(2023)Mazurowski, Dong, Gu, Yang, Konz, and Zhang}]{mazurowski2023segment}
\bibinfo{author}{M.~A. Mazurowski}, \bibinfo{author}{H.~Dong}, \bibinfo{author}{H.~Gu}, \bibinfo{author}{J.~Yang}, \bibinfo{author}{N.~Konz}, \bibinfo{author}{Y.~Zhang},
\newblock \bibinfo{title}{Segment anything model for medical image analysis: an experimental study},
\newblock \bibinfo{journal}{Medical Image Analysis} \bibinfo{volume}{89} (\bibinfo{year}{2023}) \bibinfo{pages}{102918}.
\bibitem[{Huang et~al.(2024)Huang, Yang, Liu, Zhou, Chang, Zhou, Chen, Yu, Chen, Chen et~al.}]{huang2024segment}
\bibinfo{author}{Y.~Huang}, \bibinfo{author}{X.~Yang}, \bibinfo{author}{L.~Liu}, \bibinfo{author}{H.~Zhou}, \bibinfo{author}{A.~Chang}, \bibinfo{author}{X.~Zhou}, \bibinfo{author}{R.~Chen}, \bibinfo{author}{J.~Yu}, \bibinfo{author}{J.~Chen}, \bibinfo{author}{C.~Chen}, et~al.,
\newblock \bibinfo{title}{Segment anything model for medical images?},
\newblock \bibinfo{journal}{Medical Image Analysis} \bibinfo{volume}{92} (\bibinfo{year}{2024}) \bibinfo{pages}{103061}.
\bibitem[{Zhang et~al.(2024)Zhang, Shen, and Jiao}]{zhang2024segment}
\bibinfo{author}{Y.~Zhang}, \bibinfo{author}{Z.~Shen}, \bibinfo{author}{R.~Jiao},
\newblock \bibinfo{title}{Segment anything model for medical image segmentation: Current applications and future directions},
\newblock \bibinfo{journal}{Computers in Biology and Medicine}  (\bibinfo{year}{2024}) \bibinfo{pages}{108238}.
\bibitem[{Zhang and Liu(2023)}]{zhang2023customized}
\bibinfo{author}{K.~Zhang}, \bibinfo{author}{D.~Liu},
\newblock \bibinfo{title}{Customized segment anything model for medical image segmentation},
\newblock \bibinfo{journal}{arXiv preprint arXiv:2304.13785}  (\bibinfo{year}{2023}).
\bibitem[{Zhu et~al.(2024)Zhu, Qi, and Wu}]{zhu2024medical}
\bibinfo{author}{J.~Zhu}, \bibinfo{author}{Y.~Qi}, \bibinfo{author}{J.~Wu},
\newblock \bibinfo{title}{Medical sam 2: Segment medical images as video via segment anything model 2},
\newblock \bibinfo{journal}{arXiv preprint arXiv:2408.00874}  (\bibinfo{year}{2024}).
\bibitem[{Kirillov et~al.(2023)Kirillov, Mintun, Ravi, Mao, Rolland, Gustafson, Xiao, Whitehead, Berg, Lo et~al.}]{kirillov2023segment}
\bibinfo{author}{A.~Kirillov}, \bibinfo{author}{E.~Mintun}, \bibinfo{author}{N.~Ravi}, \bibinfo{author}{H.~Mao}, \bibinfo{author}{C.~Rolland}, \bibinfo{author}{L.~Gustafson}, \bibinfo{author}{T.~Xiao}, \bibinfo{author}{S.~Whitehead}, \bibinfo{author}{A.~C. Berg}, \bibinfo{author}{W.-Y. Lo}, et~al.,
\newblock \bibinfo{title}{Segment anything},
\newblock in: \bibinfo{booktitle}{Proceedings of the IEEE/CVF International Conference on Computer Vision}, \bibinfo{year}{2023}, pp. \bibinfo{pages}{4015--4026}.
\bibitem[{Liu et~al.(2023)Liu, Shen, Pun, and Cun}]{liu2023explicit}
\bibinfo{author}{W.~Liu}, \bibinfo{author}{X.~Shen}, \bibinfo{author}{C.-M. Pun}, \bibinfo{author}{X.~Cun},
\newblock \bibinfo{title}{Explicit visual prompting for low-level structure segmentations},
\newblock in: \bibinfo{booktitle}{Proceedings of the IEEE/CVF Conference on Computer Vision and Pattern Recognition}, \bibinfo{year}{2023}, pp. \bibinfo{pages}{19434--19445}.
\bibitem[{Chen et~al.(2023)Chen, Zhu, Deng, Cao, Wang, Zhang, Li, Sun, Zang, and Mao}]{chen2023sam}
\bibinfo{author}{T.~Chen}, \bibinfo{author}{L.~Zhu}, \bibinfo{author}{C.~Deng}, \bibinfo{author}{R.~Cao}, \bibinfo{author}{Y.~Wang}, \bibinfo{author}{S.~Zhang}, \bibinfo{author}{Z.~Li}, \bibinfo{author}{L.~Sun}, \bibinfo{author}{Y.~Zang}, \bibinfo{author}{P.~Mao},
\newblock \bibinfo{title}{Sam-adapter: Adapting segment anything in underperformed scenes},
\newblock in: \bibinfo{booktitle}{Proceedings of the IEEE/CVF International Conference on Computer Vision}, \bibinfo{year}{2023}, pp. \bibinfo{pages}{3367--3375}.
\bibitem[{Liu et~al.(2022)Liu, Tam, Muqeeth, Mohta, Huang, Bansal, and Raffel}]{liu2022few}
\bibinfo{author}{H.~Liu}, \bibinfo{author}{D.~Tam}, \bibinfo{author}{M.~Muqeeth}, \bibinfo{author}{J.~Mohta}, \bibinfo{author}{T.~Huang}, \bibinfo{author}{M.~Bansal}, \bibinfo{author}{C.~A. Raffel},
\newblock \bibinfo{title}{Few-shot parameter-efficient fine-tuning is better and cheaper than in-context learning},
\newblock \bibinfo{journal}{Advances in Neural Information Processing Systems} \bibinfo{volume}{35} (\bibinfo{year}{2022}) \bibinfo{pages}{1950--1965}.
\bibitem[{Xu et~al.(2012)Xu, Zhu, Chang, and Tu}]{xu2012multiple}
\bibinfo{author}{Y.~Xu}, \bibinfo{author}{J.-Y. Zhu}, \bibinfo{author}{E.~Chang}, \bibinfo{author}{Z.~Tu},
\newblock \bibinfo{title}{Multiple clustered instance learning for histopathology cancer image classification, segmentation and clustering},
\newblock in: \bibinfo{booktitle}{2012 IEEE Conference on Computer Vision and Pattern Recognition}, \bibinfo{organization}{IEEE}, \bibinfo{year}{2012}, pp. \bibinfo{pages}{964--971}.
\bibitem[{Jia et~al.(2017)Jia, Huang, Eric, Chang, and Xu}]{jia2017constrained}
\bibinfo{author}{Z.~Jia}, \bibinfo{author}{X.~Huang}, \bibinfo{author}{I.~Eric}, \bibinfo{author}{C.~Chang}, \bibinfo{author}{Y.~Xu},
\newblock \bibinfo{title}{Constrained deep weak supervision for histopathology image segmentation},
\newblock \bibinfo{journal}{IEEE transactions on medical imaging} \bibinfo{volume}{36} (\bibinfo{year}{2017}) \bibinfo{pages}{2376--2388}.
\bibitem[{Han et~al.(2022)Han, Lin, Mai, Wang, Zhang, Zhao, Chen, Pan, Shi, Xu et~al.}]{han2022multi}
\bibinfo{author}{C.~Han}, \bibinfo{author}{J.~Lin}, \bibinfo{author}{J.~Mai}, \bibinfo{author}{Y.~Wang}, \bibinfo{author}{Q.~Zhang}, \bibinfo{author}{B.~Zhao}, \bibinfo{author}{X.~Chen}, \bibinfo{author}{X.~Pan}, \bibinfo{author}{Z.~Shi}, \bibinfo{author}{Z.~Xu}, et~al.,
\newblock \bibinfo{title}{Multi-layer pseudo-supervision for histopathology tissue semantic segmentation using patch-level classification labels},
\newblock \bibinfo{journal}{Medical Image Analysis} \bibinfo{volume}{80} (\bibinfo{year}{2022}) \bibinfo{pages}{102487}.
\bibitem[{Kweon and Yoon(2024)}]{kweon2024sam}
\bibinfo{author}{H.~Kweon}, \bibinfo{author}{K.-J. Yoon},
\newblock \bibinfo{title}{From sam to cams: Exploring segment anything model for weakly supervised semantic segmentation},
\newblock in: \bibinfo{booktitle}{Proceedings of the IEEE/CVF Conference on Computer Vision and Pattern Recognition}, \bibinfo{year}{2024}, pp. \bibinfo{pages}{19499--19509}.
\bibitem[{Ren et~al.(2024)Ren, Liu, Zeng, Lin, Li, Cao, Chen, Huang, Chen, Yan et~al.}]{ren2024grounded}
\bibinfo{author}{T.~Ren}, \bibinfo{author}{S.~Liu}, \bibinfo{author}{A.~Zeng}, \bibinfo{author}{J.~Lin}, \bibinfo{author}{K.~Li}, \bibinfo{author}{H.~Cao}, \bibinfo{author}{J.~Chen}, \bibinfo{author}{X.~Huang}, \bibinfo{author}{Y.~Chen}, \bibinfo{author}{F.~Yan}, et~al.,
\newblock \bibinfo{title}{Grounded sam: Assembling open-world models for diverse visual tasks},
\newblock \bibinfo{journal}{arXiv preprint arXiv:2401.14159}  (\bibinfo{year}{2024}).
\bibitem[{Liu et~al.(2023)Liu, Zeng, Ren, Li, Zhang, Yang, Li, Yang, Su, Zhu et~al.}]{liu2023grounding}
\bibinfo{author}{S.~Liu}, \bibinfo{author}{Z.~Zeng}, \bibinfo{author}{T.~Ren}, \bibinfo{author}{F.~Li}, \bibinfo{author}{H.~Zhang}, \bibinfo{author}{J.~Yang}, \bibinfo{author}{C.~Li}, \bibinfo{author}{J.~Yang}, \bibinfo{author}{H.~Su}, \bibinfo{author}{J.~Zhu}, et~al.,
\newblock \bibinfo{title}{Grounding dino: Marrying dino with grounded pre-training for open-set object detection},
\newblock \bibinfo{journal}{arXiv preprint arXiv:2303.05499}  (\bibinfo{year}{2023}).
\bibitem[{Yang and Gong(2024)}]{yang2024foundation}
\bibinfo{author}{X.~Yang}, \bibinfo{author}{X.~Gong},
\newblock \bibinfo{title}{Foundation model assisted weakly supervised semantic segmentation},
\newblock in: \bibinfo{booktitle}{Proceedings of the IEEE/CVF Winter Conference on Applications of Computer Vision}, \bibinfo{year}{2024}, pp. \bibinfo{pages}{523--532}.
\bibitem[{Kong et~al.(2023)Kong, Ma, Yuan, Sun, Xie, Dong, Meng, Shen, Tang, Qin et~al.}]{kong2023peeling}
\bibinfo{author}{Z.~Kong}, \bibinfo{author}{H.~Ma}, \bibinfo{author}{G.~Yuan}, \bibinfo{author}{M.~Sun}, \bibinfo{author}{Y.~Xie}, \bibinfo{author}{P.~Dong}, \bibinfo{author}{X.~Meng}, \bibinfo{author}{X.~Shen}, \bibinfo{author}{H.~Tang}, \bibinfo{author}{M.~Qin}, et~al.,
\newblock \bibinfo{title}{Peeling the onion: Hierarchical reduction of data redundancy for efficient vision transformer training},
\newblock in: \bibinfo{booktitle}{Proceedings of the AAAI Conference on Artificial Intelligence}, volume~\bibinfo{volume}{37}, \bibinfo{year}{2023}, pp. \bibinfo{pages}{8360--8368}.
\bibitem[{Zhong et~al.(2024)Zhong, Tang, He, Fang, and Yuan}]{zhong2024convolution}
\bibinfo{author}{Z.~Zhong}, \bibinfo{author}{Z.~Tang}, \bibinfo{author}{T.~He}, \bibinfo{author}{H.~Fang}, \bibinfo{author}{C.~Yuan},
\newblock \bibinfo{title}{Convolution meets lora: Parameter efficient finetuning for segment anything model},
\newblock \bibinfo{journal}{arXiv preprint arXiv:2401.17868}  (\bibinfo{year}{2024}).
\bibitem[{Gu et~al.(2024)Gu, Dong, Yang, and Mazurowski}]{gu2024build}
\bibinfo{author}{H.~Gu}, \bibinfo{author}{H.~Dong}, \bibinfo{author}{J.~Yang}, \bibinfo{author}{M.~A. Mazurowski},
\newblock \bibinfo{title}{How to build the best medical image segmentation algorithm using foundation models: a comprehensive empirical study with segment anything model},
\newblock \bibinfo{journal}{arXiv preprint arXiv:2404.09957}  (\bibinfo{year}{2024}).
\bibitem[{Yii et~al.(2023)Yii, MacGillivray, and Bernabeu}]{yii2023data}
\bibinfo{author}{F.~Yii}, \bibinfo{author}{T.~MacGillivray}, \bibinfo{author}{M.~O. Bernabeu},
\newblock \bibinfo{title}{Data efficiency of segment anything model for optic disc and cup segmentation},
\newblock in: \bibinfo{booktitle}{International conference on medical image computing and computer-assisted intervention}, \bibinfo{organization}{Springer}, \bibinfo{year}{2023}, pp. \bibinfo{pages}{336--346}.
\bibitem[{Choe and Shim(2019)}]{choe2019attention}
\bibinfo{author}{J.~Choe}, \bibinfo{author}{H.~Shim},
\newblock \bibinfo{title}{Attention-based dropout layer for weakly supervised object localization},
\newblock in: \bibinfo{booktitle}{Proceedings of the IEEE/CVF conference on computer vision and pattern recognition}, \bibinfo{year}{2019}, pp. \bibinfo{pages}{2219--2228}.
\bibitem[{Kervadec et~al.(2020)Kervadec, Dolz, Wang, Granger, and Ayed}]{kervadec2020bounding}
\bibinfo{author}{H.~Kervadec}, \bibinfo{author}{J.~Dolz}, \bibinfo{author}{S.~Wang}, \bibinfo{author}{E.~Granger}, \bibinfo{author}{I.~B. Ayed},
\newblock \bibinfo{title}{Bounding boxes for weakly supervised segmentation: Global constraints get close to full supervision},
\newblock in: \bibinfo{booktitle}{Medical imaging with deep learning}, \bibinfo{organization}{PMLR}, \bibinfo{year}{2020}, pp. \bibinfo{pages}{365--381}.
\bibitem[{Cheng et~al.(2023)Cheng, Qin, Jiang, Zhang, Lao, and Li}]{cheng2023sam}
\bibinfo{author}{D.~Cheng}, \bibinfo{author}{Z.~Qin}, \bibinfo{author}{Z.~Jiang}, \bibinfo{author}{S.~Zhang}, \bibinfo{author}{Q.~Lao}, \bibinfo{author}{K.~Li},
\newblock \bibinfo{title}{Sam on medical images: A comprehensive study on three prompt modes},
\newblock \bibinfo{journal}{arXiv preprint arXiv:2305.00035}  (\bibinfo{year}{2023}).
\bibitem[{Gaus et~al.(2024)Gaus, Bhowmik, Isaac-Medina, and Breckon}]{gaus2024performance}
\bibinfo{author}{Y.~F.~A. Gaus}, \bibinfo{author}{N.~Bhowmik}, \bibinfo{author}{B.~K. Isaac-Medina}, \bibinfo{author}{T.~P. Breckon},
\newblock \bibinfo{title}{Performance evaluation of segment anything model with variational prompting for application to non-visible spectrum imagery},
\newblock in: \bibinfo{booktitle}{Proceedings of the IEEE/CVF Conference on Computer Vision and Pattern Recognition}, \bibinfo{year}{2024}, pp. \bibinfo{pages}{3142--3152}.
\bibitem[{Cheng et~al.(2023)Cheng, Ye, Deng, Chen, Li, Wang, Su, Huang, Chen, Jiang et~al.}]{cheng2023sam2d}
\bibinfo{author}{J.~Cheng}, \bibinfo{author}{J.~Ye}, \bibinfo{author}{Z.~Deng}, \bibinfo{author}{J.~Chen}, \bibinfo{author}{T.~Li}, \bibinfo{author}{H.~Wang}, \bibinfo{author}{Y.~Su}, \bibinfo{author}{Z.~Huang}, \bibinfo{author}{J.~Chen}, \bibinfo{author}{L.~Jiang}, et~al.,
\newblock \bibinfo{title}{Sam-med2d},
\newblock \bibinfo{journal}{arXiv preprint arXiv:2308.16184}  (\bibinfo{year}{2023}).
\bibitem[{Dai et~al.(2023)Dai, Ma, Yan, Liu, Shi, Li, Shu, Wei, Zhao, Wu et~al.}]{dai2023samaug}
\bibinfo{author}{H.~Dai}, \bibinfo{author}{C.~Ma}, \bibinfo{author}{Z.~Yan}, \bibinfo{author}{Z.~Liu}, \bibinfo{author}{E.~Shi}, \bibinfo{author}{Y.~Li}, \bibinfo{author}{P.~Shu}, \bibinfo{author}{X.~Wei}, \bibinfo{author}{L.~Zhao}, \bibinfo{author}{Z.~Wu}, et~al.,
\newblock \bibinfo{title}{Samaug: Point prompt augmentation for segment anything model},
\newblock \bibinfo{journal}{arXiv preprint arXiv:2307.01187}  (\bibinfo{year}{2023}).
\bibitem[{Hu et~al.(2023)Hu, Xu, and Shi}]{hu2023efficiently}
\bibinfo{author}{X.~Hu}, \bibinfo{author}{X.~Xu}, \bibinfo{author}{Y.~Shi},
\newblock \bibinfo{title}{How to efficiently adapt large segmentation model (sam) to medical images},
\newblock \bibinfo{journal}{arXiv preprint arXiv:2306.13731}  (\bibinfo{year}{2023}).
\bibitem[{Chen et~al.(2023)Chen, Mai, Li, and Chao}]{chen2023segment}
\bibinfo{author}{T.~Chen}, \bibinfo{author}{Z.~Mai}, \bibinfo{author}{R.~Li}, \bibinfo{author}{W.-l. Chao},
\newblock \bibinfo{title}{Segment anything model (sam) enhanced pseudo labels for weakly supervised semantic segmentation},
\newblock \bibinfo{journal}{arXiv preprint arXiv:2305.05803}  (\bibinfo{year}{2023}).
\bibitem[{Wang et~al.(2023)Wang, Zhang, and Shi}]{wang2023cs}
\bibinfo{author}{L.~Wang}, \bibinfo{author}{M.~Zhang}, \bibinfo{author}{W.~Shi},
\newblock \bibinfo{title}{Cs-wscdnet: Class activation mapping and segment anything model-based framework for weakly supervised change detection},
\newblock \bibinfo{journal}{IEEE Transactions on Geoscience and Remote Sensing}  (\bibinfo{year}{2023}).
\bibitem[{Yang et~al.(2024)Yang, He, Yin, Wang, Zhang, Long, and Peng}]{yang2024weakly}
\bibinfo{author}{R.~Yang}, \bibinfo{author}{G.~He}, \bibinfo{author}{R.~Yin}, \bibinfo{author}{G.~Wang}, \bibinfo{author}{Z.~Zhang}, \bibinfo{author}{T.~Long}, \bibinfo{author}{Y.~Peng},
\newblock \bibinfo{title}{Weakly-semi supervised extraction of rooftop photovoltaics from high-resolution images based on segment anything model and class activation map},
\newblock \bibinfo{journal}{Applied Energy} \bibinfo{volume}{361} (\bibinfo{year}{2024}) \bibinfo{pages}{122964}.
\bibitem[{Ma et~al.(2024)Ma, He, Li, Han, You, and Wang}]{ma2024segment}
\bibinfo{author}{J.~Ma}, \bibinfo{author}{Y.~He}, \bibinfo{author}{F.~Li}, \bibinfo{author}{L.~Han}, \bibinfo{author}{C.~You}, \bibinfo{author}{B.~Wang},
\newblock \bibinfo{title}{Segment anything in medical images},
\newblock \bibinfo{journal}{Nature Communications} \bibinfo{volume}{15} (\bibinfo{year}{2024}) \bibinfo{pages}{654}.
\bibitem[{Selvaraju et~al.(2020)Selvaraju, Cogswell, Das, Vedantam, Parikh, and Batra}]{selvaraju2020grad}
\bibinfo{author}{R.~R. Selvaraju}, \bibinfo{author}{M.~Cogswell}, \bibinfo{author}{A.~Das}, \bibinfo{author}{R.~Vedantam}, \bibinfo{author}{D.~Parikh}, \bibinfo{author}{D.~Batra},
\newblock \bibinfo{title}{Grad-cam: visual explanations from deep networks via gradient-based localization},
\newblock \bibinfo{journal}{International journal of computer vision} \bibinfo{volume}{128} (\bibinfo{year}{2020}) \bibinfo{pages}{336--359}.
\bibitem[{Chattopadhay et~al.(2018)Chattopadhay, Sarkar, Howlader, and Balasubramanian}]{chattopadhay2018grad}
\bibinfo{author}{A.~Chattopadhay}, \bibinfo{author}{A.~Sarkar}, \bibinfo{author}{P.~Howlader}, \bibinfo{author}{V.~N. Balasubramanian},
\newblock \bibinfo{title}{Grad-cam++: Generalized gradient-based visual explanations for deep convolutional networks},
\newblock in: \bibinfo{booktitle}{2018 IEEE winter conference on applications of computer vision (WACV)}, \bibinfo{organization}{IEEE}, \bibinfo{year}{2018}, pp. \bibinfo{pages}{839--847}.
\bibitem[{Muhammad and Yeasin(2020)}]{muhammad2020eigen}
\bibinfo{author}{M.~B. Muhammad}, \bibinfo{author}{M.~Yeasin},
\newblock \bibinfo{title}{Eigen-cam: Class activation map using principal components},
\newblock in: \bibinfo{booktitle}{2020 international joint conference on neural networks (IJCNN)}, \bibinfo{organization}{IEEE}, \bibinfo{year}{2020}, pp. \bibinfo{pages}{1--7}.
\bibitem[{Qian et~al.(2022)Qian, Li, Lai, Chang, Wei, Fan, and Xu}]{qian2022transformer}
\bibinfo{author}{Z.~Qian}, \bibinfo{author}{K.~Li}, \bibinfo{author}{M.~Lai}, \bibinfo{author}{E.~I.-C. Chang}, \bibinfo{author}{B.~Wei}, \bibinfo{author}{Y.~Fan}, \bibinfo{author}{Y.~Xu},
\newblock \bibinfo{title}{Transformer based multiple instance learning for weakly supervised histopathology image segmentation},
\newblock in: \bibinfo{booktitle}{International Conference on Medical Image Computing and Computer-Assisted Intervention}, \bibinfo{organization}{Springer}, \bibinfo{year}{2022}, pp. \bibinfo{pages}{160--170}.
\bibitem[{Li et~al.(2023)Li, Qian, Han, Eric, Chang, Wei, Lai, Liao, Fan, and Xu}]{li2023weakly}
\bibinfo{author}{K.~Li}, \bibinfo{author}{Z.~Qian}, \bibinfo{author}{Y.~Han}, \bibinfo{author}{I.~Eric}, \bibinfo{author}{C.~Chang}, \bibinfo{author}{B.~Wei}, \bibinfo{author}{M.~Lai}, \bibinfo{author}{J.~Liao}, \bibinfo{author}{Y.~Fan}, \bibinfo{author}{Y.~Xu},
\newblock \bibinfo{title}{Weakly supervised histopathology image segmentation with self-attention},
\newblock \bibinfo{journal}{Medical Image Analysis} \bibinfo{volume}{86} (\bibinfo{year}{2023}) \bibinfo{pages}{102791}.

\end{thebibliography}
\end{document}